\documentclass[runningheads]{llncs}

\usepackage[T1]{fontenc}
\usepackage{bbm}
\usepackage{amsmath}
\usepackage{amssymb}
\usepackage{booktabs}
\usepackage{array}
\usepackage{graphicx}
\usepackage{siunitx}
\usepackage[hidelinks]{hyperref}

\newcommand{\NA}{---}

\begin{document}

\title{Stroke outcome and evolution prediction from CT brain using a spatiotemporal diffusion autoencoder}
\titlerunning{Stroke outcome prediction using a diffusion autoencoder}

\author{Adam Marcus\inst{1}\orcidID{0000-0003-1687-4549} \and
Paul Bentley\inst{1}\orcidID{0000-0001-8036-7010} \and
Daniel Rueckert\inst{1,2}\orcidID{0000-0002-5683-5889}}

\authorrunning{A. Marcus et al.}

\institute{Imperial College London, London, UK  \email{\{adam.marcus11,p.bentley,d.rueckert\}@imperial.ac.uk} \\ \and
Technische Universität München, München, Germany\\
\email{daniel.rueckert@tum.de}}

\maketitle 

\begin{abstract}

Stroke is a major cause of death and disability worldwide. Accurate outcome and evolution prediction has the potential to revolutionize stroke care by individualizing clinical decision-making leading to better outcomes. However, despite a plethora of attempts and the rich data provided by neuroimaging, modelling the ultimate fate of brain tissue remains a challenging task. In this work, we apply recent ideas in the field of diffusion probabilistic models to generate a self-supervised semantically meaningful stroke representation from Computed Tomography (CT) images. We then improve this representation by extending the method to accommodate longitudinal images and the time from stroke onset. The effectiveness of our approach is evaluated on a dataset consisting of 5,824 CT images from 3,573 patients across two medical centers with minimal labels. Comparative experiments show that our method achieves the best performance for predicting next-day severity and functional outcome at discharge.

\end{abstract}

\section{Introduction}

Stroke is a major global health problem \cite{who_2018}. It often begins as the result of impaired blood flow in the brain due to a blood clot. As the brain becomes damaged, it swells, which leads to visible changes in Computed Tomography (CT) scans. Accordingly, imaging plays an essential role, and stroke patients frequently have multiple scans throughout their recovery. The impact of a stroke is usually significant and ranges from reduced health-related quality of life to disability and death. Accurately predicting these consequences and how the disease evolves could revolutionize stroke management and usher in a new age of precision medicine. In this era, aftercare decisions could be optimal, and personalized treatment choices, along with patient-specific rehabilitation targets, could lead to better outcomes \cite{bonkhoff2022precision}. Despite numerous attempts, however, realizing this goal remains far. 

\subsubsection{Related Work}

Current approaches have focused on predicting the functional outcome of treatment and, to a lesser extent, stroke severity. Functional outcome is generally measured using the modified Rankin Scale (mRS) \cite{van1988interobserver}, that ranges from 0 (no disability) to 6 (death) and can be dichotomized into independent ($<3$) or requires assistance ($\geq 3$). Severity is often assessed with the National Institutes of Health Stroke Scale (NIHSS) \cite{brott1989measurements}, that ranges from 0 (no symptoms) to 42 (most severe neurologic deficit). The majority of studies have exclusively used clinical information to predict these outcomes with methods such as logistic regression \cite{heo2018machine,venema2017selection}, support vector machines \cite{bentley2014prediction}, random forests \cite{van2018predicting}, and artificial neural networks (ANN) \cite{asadi2014machine,van2018predicting}. By comparison, fewer studies have attempted to utilize imaging data. Notably, Bacchi et al. \cite{bacchi2020deep} proposed using a 3D convolutional neural network (CNN) based on the VGG architecture \cite{simonyan2014very} to process non-contrast CT (NCCT) combined with an ANN. They achieved an area under the receiver operator characteristic curve (AUC) for next-day improvement in severity classification of 0.70. Nawabi et al. \cite{nawabi2021imaging} applied a random forest model to radiomic features derived from NCCT and attained an AUC for dichotomized mRS at discharge of 0.80. A number of studies have also applied convolutional models to predict outcomes using other modalities, such as magnetic resonance imaging (MRI) \cite{bourached2022scaling}. These works have all relied on supervised learning techniques, thereby overlooking the potential benefit of unlabeled images, often more readily available in the medical domain.

The scarcity of labeled images, particularly in medicine, has spurred significant interest self-supervised learning (SSL). In SSL, methods can be broadly categorized into generative and discriminative. Generative methods typically involve an autoencoder that attempts to learn a compressed representation of its input data. Discriminative methods involve solving a task by learning a decision boundary through its data. For vision problems, discriminative techniques are currently considered the more performant. Specifically, joint-embedding approaches, such as contrastive learning \cite{chen2020simple}, that aims to align the embedded representations of augmented views of the same image. However, recent developments enabling diffusion models, which have achieved remarkable performance in generative tasks, to be used as an autoencoder may change this \cite{preechakul2022diffusion}. 

\subsubsection{Contributions}

In this work, our objective is to derive an imaging-based feature representation that faithfully captures the entire stroke trajectory from onset to recovery. Our method utilizes diffusion probabilistic models and is motivated by recent developments that allow their use as an autoencoder; the hope is that their higher fidelity image reconstruction translates to a better representation for outcome prediction. The main contributions are: (1) We apply diffusion probabilistic models to generate a self-supervised semantically meaningful stroke representation. (2) We then improve this representation by extending the method to accommodate longitudinal images and the time from stroke onset. (3) We evaluate the effectiveness of these representations when applied to baseline images to predict valuable stroke outcomes, namely, next-day severity and functional status at discharge.

\section{Method}

\begin{figure}
\includegraphics[width=\textwidth]{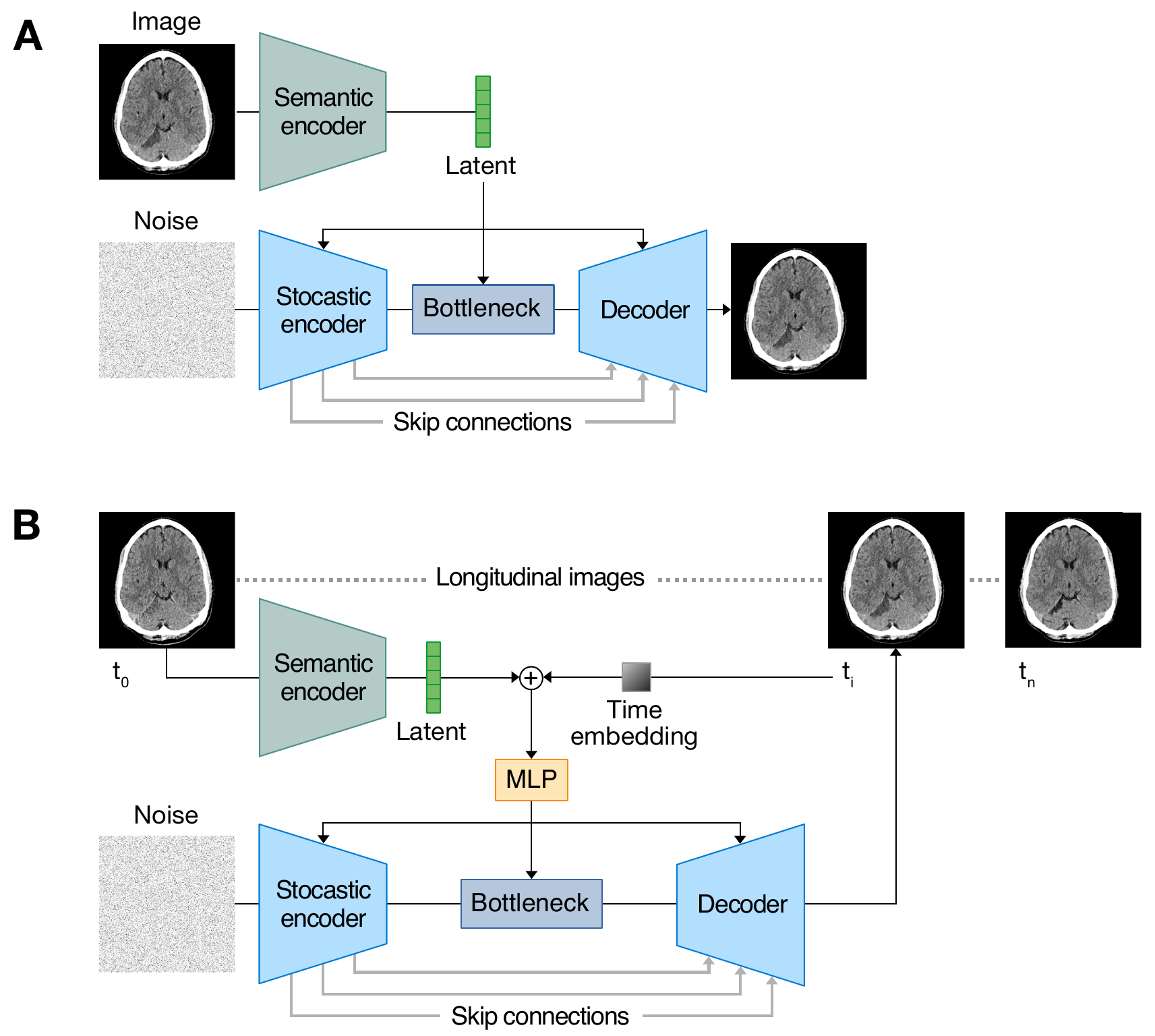}
\caption{Overview of the (A) spatial and (B) spatiotemporal diffusion autoencoder approaches. In both cases, a semantic encoder takes an image containing a stroke lesion and maps it to a latent code. A Denoising Diffusion Probabilistic Models (DDPM) \cite{ho2020denoising} is then conditioned on this latent code to denoise a different image of the same lesion taken either at (A) the same time or (B) a future point in time. For the spatiotemporal method, the latent code is also concatenated with the future time using a multilayer perceptron (MLP). After training, the semantic encoder can then be fine-tuned with minimal data and used to predict a stroke outcome.} 
\label{fig:architecture}
\end{figure}

Our spatial method is based on the recently proposed diffusion autoencoder \cite{preechakul2022diffusion}, which we then extend to incorporate longitudinal images as our spatiotemporal approach. An overview of these approaches can be seen in \autoref{fig:architecture}. Central to both of our methods is the use of Denoising Diffusion Probabilistic Models (DDPM) \cite{ho2020denoising} that model a distribution of images by learning a denoising process. A successful process can predict the varying amount of Gaussian noise, $\epsilon \sim \mathcal{N}(\textbf{0}, \textbf{I})$ with respect to timesteps $t$ (out of $T$), added to a clean image $\mathbf{x}_0$, by learning a function $\epsilon_\theta(\mathbf{x}_t, t)$ that takes the noisy image $\mathbf{x}_t$ as input. Here $\epsilon_\theta(\cdot)$ is often modeled as a U-Net with parameters $\theta$ and a loss function $||\epsilon_\theta(\mathbf{x}_t, t) - \epsilon||$.

\begin{table}
    \caption{Network architecture of our diffusion model based on the improved Denoising Diffusion Probabilistic Model (DDPM) of Dhariwal et al. \cite{dhariwal2021diffusion}.}
    \label{tab:architecture}
    \centering
    \begin{tabular*}{0.65\linewidth}{@{\extracolsep{\fill}}lc}
    \toprule
        Parameter & Value \\
    \midrule
        Base channels & 16 \\
        Channel multipliers & [1, 2, 4, 8, 16, 32, 64] \\
        Attention resolution & [16] \\
        Encoder base channels & 16 \\
        Encoder attention resolution & [16] \\
        Encoder channel multipliers & [1, 2, 4, 8, 16, 32, 64] \\
        $z$ size & 512 \\
        $\beta$ scheduler & Linear \\
        Training $T$ & 1000 \\
    \bottomrule
    \end{tabular*}
\end{table}
\subsubsection{Spatial DDPM} 

Our goal is to train a semantic encoder $\text{Enc}(\textbf{x})$ able to produce a meaningful stroke representation $\textbf{z}$. Ideally, this representation should be invariant to changes in view and other low-level variations in the image. We therefore design a framework that learns this semantic latent code along with a noisy latent code which results from using a DDPM. This is achieved through the use of a DDPM conditioned on the semantic latent code. In our experiments we employ a ResNet-50 model \cite{he2016deep} as the semantic encoder to produce a vector of dimension $d = 512$, which was chosen to resemble the style vector in StyleGAN \cite{karras2019style}. 
During training, a pair of 2D image slices ($\mathbf{x}_a$, $\mathbf{x}_b$) showing the same stroke lesion are sampled and each slice is augmented with a CT-specific strategy including: random axial plane flips; \SI{+-5}{\percent} isotropic scaling; \SI{+-20}{\milli\metre} translation; and \SI{+-0.5}{\radian} axial rotation. These images are then used as inputs to the semantic encoder and DDPM which are jointly optimized with a revised $\mathcal{L}_{simple}$ loss \cite{ho2020denoising}:

\begin{equation}
\begin{gathered}
z = \text{Enc}(\textbf{x}_a) \\
\mathcal{L}_{simple} = \sum_{t=1}^T \mathbb{E}_{\mathbf{x}_a,\mathbf{x}_b,\epsilon_t} \left[||\epsilon_\theta(\mathbf{x}_{b_t}, t, z) - \epsilon_t||_2^2\right]
\end{gathered}
\end{equation}

Here $\epsilon_t \in \mathbb{R}^{h \times w} \sim \mathcal{N}(\textbf{0}, \textbf{I})$, $h \times w$ is the spatial resolution, $\mathbf{x}_{b_t}=\sqrt{\alpha_t}\mathbf{x}_b+\sqrt{1-\alpha_t}\epsilon_t$, $T = 1000$ or an equally large number, and $\alpha$ relates to the variance schedule of the Guassian noise \cite{ho2020denoising}. Our conditional DDPM is implemented as a modified U-Net \cite{salimans2016improved} with the improvements by Dhariwal et al. \cite{dhariwal2021diffusion} that include increasing the attention channels and using BigGAN \cite{brock2018large} residual blocks for up and downsampling. We make two additional architectural changes with the other hyperparameters specified in \autoref{tab:architecture}. First, we introduce group normalization (GroupNorm) \cite{wu2018group} layers after every 1×1 convolution, acting as skip connections in each residual block. Empirically we found this stabilizes training and is particularly effective when using a deeper network with more blocks. Second, we replace the other GroupNorm layers throughout the network with Adaptive Spatial Group Normalization (AdaSpaGN) that scales and shifts the normalized feature maps. This is defined where $k \in \mathbb{R}^{c \times h \times w}$ is the feature maps with channels $c$ obtained from the U-Net, diffusion timesteps is $t$, the semantic latent code is $z$, $\psi$ is a sinusoidal encoding function \cite{vaswani2017attention} and MLP is a multilayered perceptron:

\begin{equation}
\begin{gathered}
(s, b) \in \mathbb{R}^{2 \times c} = \text{MLP}(z, \psi(t)) \\
\text{AdaSpaGN}(k, s, b) = s \text{GroupNorm}(k) + b
\end{gathered}
\end{equation}

\subsubsection{Spatiotemporal DDPM}

To make the representation produced by the semantic encoder more suited for clinical outcome prediction, our key insight is that it should be invariant to the point in time that the image was taken along the stroke trajectory. We, therefore, extend our spatial method by sampling a different pair of images during training showing the same lesion but now at distinct points in time. The semantic encoder receives $\mathbf{x}_a$, which is always a view from the earliest scan, and the DDPM receives a noisy variant of $\mathbf{x}_b$ taken at a later time. In our ablation study, we explore whether it is necessary to enforce this ordering.

Intending to reduce the difficulty of learning this new task, we also condition the DDPM on the time image $\mathbf{x}_b$ was taken since stroke onset. We log-transform these times to account for a skewed distribution and replace the AdaSpaGN layers with Adaptive Temporal Group Normalization (AdaTempGN) which is defined where time is $n$ as:

\begin{equation}
\begin{gathered}
(s, b) \in \mathbb{R}^{2 \times c} = \text{MLP}(z, \psi(\log(n)), \psi(t)) \\
\text{AdaTempGN}(k, s, b) = s \text{GroupNorm}(k) + b
\end{gathered}
\end{equation}

\section{Experiments}

\subsection{Materials}

\subsubsection{Dataset}

Experiments were performed on a dataset of 3,573 acute ischemic stroke patients collected across two clinical sites from 2010 to 2019. Patients were divided so that a fixed random 20\% split were used for testing and the remainder for training and validation using a five-fold cross-validation approach. \autoref{tab:dataset-demographics} details the characteristics of these groups. A high number of patients were missing outcome measures but contained timing information. Muschelli's \cite{muschelli2019recommendations} recommended pipeline was used to extract and anonymize the patients' non-contrast CT images. Full ethical approval was granted by Wales REC 3 reference number 16/WA/0361.

\begin{table}
    \caption{Population characteristics of the clinical dataset.}
    \label{tab:dataset-demographics}
    \centering
    \begin{tabular*}{\linewidth}{@{\extracolsep{\fill}}m{5cm}cc}
    \toprule
        \multicolumn{1}{c}{Characteristic} & \multicolumn{1}{m{3.75cm}}{\centering Train and validation set (n = 2858)} & 
        Test set (n = 715) \\
    \midrule
        Number of CT, mean (range) & 1.63 (1--5) & 1.63 (1--5) \\
        \addlinespace[0.25em]
        Missing outcome, n (\%) & 1178 (41.2\%) & 276 (38.6\%) \\
        \addlinespace[0.25em]
        Age (years), median (IQR) & 75.0 (62.4--83.7) & 75.0 (65.0--83.0) \\
        \addlinespace[0.25em]
        Female sex, n (\%) & 914 (54.4\%) & 242 (55.1\%) \\
        \addlinespace[0.25em]
        ASPECTS, median (IQR) & 10 (9--10) & 10 (9--10) \\
        \addlinespace[0.25em]
        NIHSS on admission, median (IQR) & 7 (4--13) & 7 (4--13) \\
        \addlinespace[0.25em]
        NIHSS at 24 hours, median (IQR) & 5 (2--11) & 5 (2--12) \\
        \addlinespace[0.25em]
        mRS on discharge, median (IQR) & 3 (1--4) & 3 (1--4) \\
        \addlinespace[0.25em]
        Time from symptom onset to first CT (minutes), median (IQR) & 180 (95--522) & 169 (90--550) \\
    \bottomrule
\end{tabular*}
\begin{tabular}{p{\dimexpr\linewidth-2\tabcolsep\relax}}
    \addlinespace[0.5em]
    \begin{footnotesize}
        IQR = Interquartile range; ASPECTS = Alberta stroke program early CT score; NIHSS = National Institutes of Health Stroke Scale \cite{brott1989measurements}; mRS = Modified Rankin score \cite{van1988interobserver}
    \end{footnotesize}
\end{tabular}
\end{table}

\subsubsection{Evaluation}

A positive stroke outcome was defined as either a mRS $<3$ or the improvement in next-day NIHSS of $\geq 4$ points. These thresholds were selected due to their use in previous studies \cite{hacke2008thrombolysis,bacchi2020deep,samak2022fema}. Classification performance was evaluated using accuracy (ACC), F1 score, and AUC. To determine significant differences in AUC, permutation testing was used \cite{bandos2005permutation}. The Fréchet inception distance (FID) \cite{heusel2017gans} and mean squared error (MSE) were used to assess the quality of image reconstruction.

\subsubsection{Implementation}

All models were implemented using PyTorch version 1.13 on a machine with 3.80GHz Intel\textsuperscript{\textregistered} Core\textsuperscript{TM} i7-10700K CPU and an NVIDIA GeForce RTX 3080 10GB GPU. The AdamW \cite{loshchilov2017decoupled} optimizer was used with a learning rate of $10^{-3}$ when training from scratch and $10^{-4}$ when fine tuning with a weight decay coefficient of $10^{-2}$. To ensure fairness, all models were trained initially for 1 million steps and a ResNet-50 \cite{he2016deep} was used as the semantic encoder. If required, the encoder was then independently fine-tuned for each prediction task, for an additional 100,000 steps, with all but the final layer weights frozen for the first 10,000 steps. Lesion containing slices were identified using a previously described model \cite{marcus2022concurrent} and linearly sampled from the original volumes to a uniform size of 512 × 512 with a spatial resolution of 0.45 × 0.45mm$^2$. Images were clipped based on the 0.5 and 99.5th percentile then normalized using Z-score. Final inference of the models required approximately a second per subject.

\subsection{Results}

\begin{table}
    \caption{Stroke outcome prediction results obtained by our method and spatiotemporal ablation variants compared to baseline approaches. In our method, a pair of 2D image slices, $\mathbf{x}_a$ and $\mathbf{x}_b$, showing the same stroke lesion are sampled, where $\mathbf{x}_a$ is always from the earliest scan and the DDPM receives a noisy variant of $\mathbf{x}_b$ taken at a later time. In the \textit{Any forward pair} variant, $\mathbf{x}_a$ is from any time prior to $\mathbf{x}_b$, and in the \textit{Any pair} variant, $\mathbf{x}_a$ and $\mathbf{x}_b$ can be views from any time. For fair comparison all methods utilized a ResNet-50 \cite{he2016deep} as the semantic encoder and followed an identical fine-tuning procedure, unless trained directly.}
    \label{tab:results}
    \centering
    \begin{tabular*}{\linewidth}{@{\extracolsep{\fill}}rrccccccccc}
    \toprule
        & & \multicolumn{3}{c}{24-hour NIHSS} & \multicolumn{3}{c}{Discharge mRS} & \\
        \cmidrule(lr){3-5} \cmidrule(lr){6-8}
        \multicolumn{2}{c}{Method} & AUC & ACC & F1 & AUC & ACC & F1 & FID & MSE \\
    \midrule
        CNN & Direct training & 0.584 & 59.9 & 52.0 & 0.702 & 60.4 & 57.4 & \NA & \NA \\
        & VICReg \cite{bardes2021vicreg} & 0.582 & 59.9 & 52.0 & 0.711 & 63.1 & 58.6 & \NA & \NA \\ [0.25em]
        Autoencoder & Variational \cite{kingma2013auto} & 0.628 & 62.9 & 63.6 & 0.726 & 64.4 & 63.5 & 8.9 & 69.9 \\
        & Diffusion \cite{preechakul2022diffusion} & 0.623 & 62.6 & 64.2 & 0.735 & 65.8 & 64.2 & \bfseries 7.1 & \bfseries 47.7 \\ [0.25em]
        Ours & Spatial & 0.648 & 63.6 & 65.4 & 0.757 & 67.7 & 69.2 & 7.3 & 47.9 \\
        & Spatiotemporal & \bfseries 0.669 & 63.6 & 67.4 & 0.788 & \bfseries 70.8 & 71.6 & 7.4 & 48.2 \\
    \cmidrule(lr){2-2}
        & No augmentation & 0.666 & 63.6 & 67.4 & 0.785 & 70.2 & 69.8 & 7.4 & 47.9 \\ 
        & Any forward pair & 0.663 & 63.3 & \bfseries 67.6 & \bfseries 0.789 & 70.6 & 71.1 & 7.3 & 48.1 \\ 
        & Any pair & 0.667 & \bfseries 63.8 & 67.2 & 0.787 & 70.8 & \bfseries 72.0 & 7.3 & 48.2 \\ 
    \bottomrule
\end{tabular*}
\begin{tabular}{p{\dimexpr\linewidth-2\tabcolsep\relax}}
    \addlinespace[0.5em]
    \begin{footnotesize}
        NIHSS = National Institutes of Health Stroke Scale \cite{brott1989measurements}; mRS = Modified Rankin score \cite{van1988interobserver}; AUC = Area under the receiver operator characteristic curve; ACC = Accuracy; FID = Fréchet inception distance \cite{heusel2017gans}; MSE = Mean squared error; CNN = Convolutional Neural Network
    \end{footnotesize}
\end{tabular}
\end{table}

\begin{figure}
\centering
\includegraphics[width=0.8\textwidth]{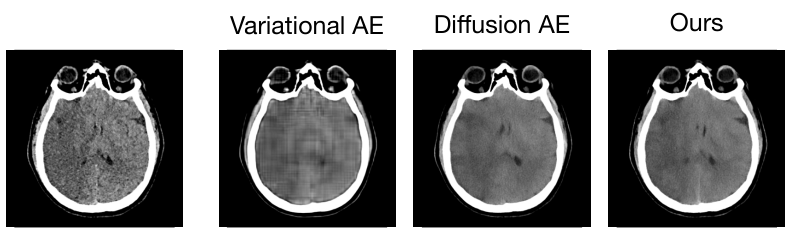}
\caption{Example reconstructed image of a right middle cerebral artery (MCA) stroke from our test set for different methods. The performance of our spatiotemporal approach and the diffusion autoencoder (AE) are similar and superior to a variational AE.} 
\label{fig:example}
\end{figure}

\subsubsection{Comparison with Baseline}

The quantitative results can be seen in \autoref{tab:results}. First, we recognize that across all the models tested, predicting improvement in severity over 24 hours appears to be a more challenging task than functional status at discharge. It seems plausible that this may be due to the models not knowing the given treatment, which is likely more impactful in the short term. Second, both our method and autoencoders appear to increase performance over training directly from the labeled images. This is also true when employing non-contrastive pre-training with VICReg \cite{bardes2021vicreg}, which was selected over other methods due to its robustness to smaller batch sizes. Third, we note that the diffusion-based models provide superior image reconstruction ability, which is further supported by qualitative evaluation seen in \autoref{fig:example}. Fourth, despite offering the best prediction performance, our approach slightly underperforms in image reconstruction. This suggests that subtle local image features, which global image metrics may not fully capture, hold greater significance in outcome prediction. Finally, we observe that our spatiotemporal approach shows significantly ($p$ value $\leq$0.05) greater AUC for predicting discharge mRS over other methods.

\subsubsection{Comparison with the Literature}

There are few existing studies to which we can fairly compare our results. Partly this is due to most studies focusing on predicting the justifiably more clinically relevant 90-day rather than discharge mRS. This is a limitation of the current work, as the dataset used did not contain longer-term measures. An issue further magnified by the lack of a readily available public dataset. However, it should be noted that both next-day NIHSS and discharge mRS are highly associated with 90-day mRS \cite{hendrix2022nihss,elhabr2021predicting}. In comparison with Bacchi et al. \cite{bacchi2020deep}, who also predicted an improvement in 24-hour NIHSS, we achieved a higher AUC of 0.669 compared to 0.63 using only imaging features. Similarly, we attained a comparable AUC for predicting dichotomized discharge mRS of 0.789 to 0.80 by Nawabi et al. \cite{nawabi2021imaging}. Although noteworthy that they looked at hemorrhagic rather than ischemic strokes.

\subsubsection{Ablation Study}

To justify our design decisions and verify the effectiveness of our approach, we conducted several ablation experiments, shown in \autoref{tab:results}. We first note that our augmentation strategy has a minimal impact on the final stroke representation leading to a similar performance in outcome prediction. Similarly, we observed no significant effect when changing the combination of images used during training. We hypothesize this may be partly due to our dataset often containing only a single pair of images per subject.

\section{Conclusion}

In this paper, we have developed an imaging-based stroke representation capturing features predictive of future events through the use of diffusion models. Our method can utilize unlabeled longitudinal images with any duration between scans. Empirical results suggest our approach offers promising performance at estimating next-day severity and functional status at discharge. Future research includes integrating clinical information and prospective validation with longer-term outcomes. We hope that this work ultimately leads to more effective personalized treatment strategies for stroke patients. 

\section{Acknowledgements}

This work was supported by the UK Research and Innovation: UKRI Center for Doctoral Training in AI for Healthcare under Grant EP/S023283/1 and UK National Institute for Health Research i4i Program under Grant II-LA-0814--20007.

\bibliographystyle{splncs04}
\bibliography{bibliography}

\end{document}